\documentclass{article}
\usepackage{ijcai22}

\usepackage{times}

\usepackage{soul}
\usepackage{url}
\usepackage[hidelinks]{hyperref}
\usepackage[utf8]{inputenc}
\usepackage[small]{caption}
\usepackage{graphicx}
\usepackage{amsmath}
\usepackage{booktabs}
\usepackage{lipsum}
\usepackage{multirow}
\usepackage{float}
\usepackage{flafter} 
\usepackage{amsmath,amssymb,booktabs,tabulary,multirow,overpic,xcolor}
\usepackage[caption=false]{subfig}
\usepackage{arydshln}

\title{MotionMixer: MLP-based 3D Human Body Pose Forecasting}

\author{
Arij Bouazizi$^{1,2}$\footnote{Contact Author}\and
Adrian Holzbock$^2$\and
Ulrich Kressel $^{1}$\and
Klaus Dietmayer$^2$\And
\\ Vasileios Belagiannis$^3$ \footnote{Most of this work was done while Vasileios Belagiannis was with Ulm University.} \\
\affiliations
$^1$Mercedes-Benz AG, Stuttgart, Germany\\
$^2$Ulm University, Ulm, Germany\\
$^3$Otto von Guericke University Magdeburg, Magdeburg, Germany\\
\emails
\{arij.bouazizi, ulrich.kressel\}@mercedes-benz.com,
\{adrian.holzbock, klaus.dietmayer\}@uni-ulm.de,
vasileios.belagiannis@ovgu.de
}

\begin{document}

\maketitle

\begin{abstract}
In this work, we present \textit{MotionMixer}, an efficient 3D human body pose forecasting model based solely on multi-layer perceptrons (MLPs). \textit{MotionMixer} learns the spatial-temporal 3D body pose dependencies by sequentially mixing both modalities. Given a stacked sequence of 3D body poses, a spatial-MLP extracts fine-grained spatial dependencies of the body joints. The interaction of the body joints over time is then modelled by a temporal MLP. The spatial-temporal mixed features are finally aggregated and decoded to obtain the future motion. To calibrate the influence of each time step in the pose sequence, we  make use of squeeze-and-excitation (SE) blocks. We evaluate our approach on Human3.6M, AMASS, and 3DPW datasets using the standard evaluation protocols. For all evaluations, we demonstrate state-of-the-art performance, while having a model with a smaller number of parameters.  Our code is available at: \url{https://github.com/MotionMLP/MotionMixer}.
\end{abstract}

\section{Introduction}
Forecasting 3D human motion is at the core of many different applications ranging from virtual reality to autonomous driving \cite {wiederer2020traffic} and robotics \cite{gui2018teaching}. Fundamentally, the task of human motion prediction is defined as the prediction of future body poses from past ones. To model the human spatial-temporal dynamics, classic approaches adopted Hidden Markov Models \cite{kulic2012incremental} or Gaussian Processes \cite{wang2007gaussian}. Despite the tangible progress in predicting periodic motion, these approaches impose strong assumptions on body pose, leading to performance degradation. Furthermore, it is difficult for these methods to forecast reliable 3D poses because of  the complicated bio-mechanical kinematics. Even manually imposing expert knowledge does not help prior-based approaches to generalise to new environments and subjects.

\begin{figure} [ht!]
    \centering
    \includegraphics[width=0.48\textwidth]{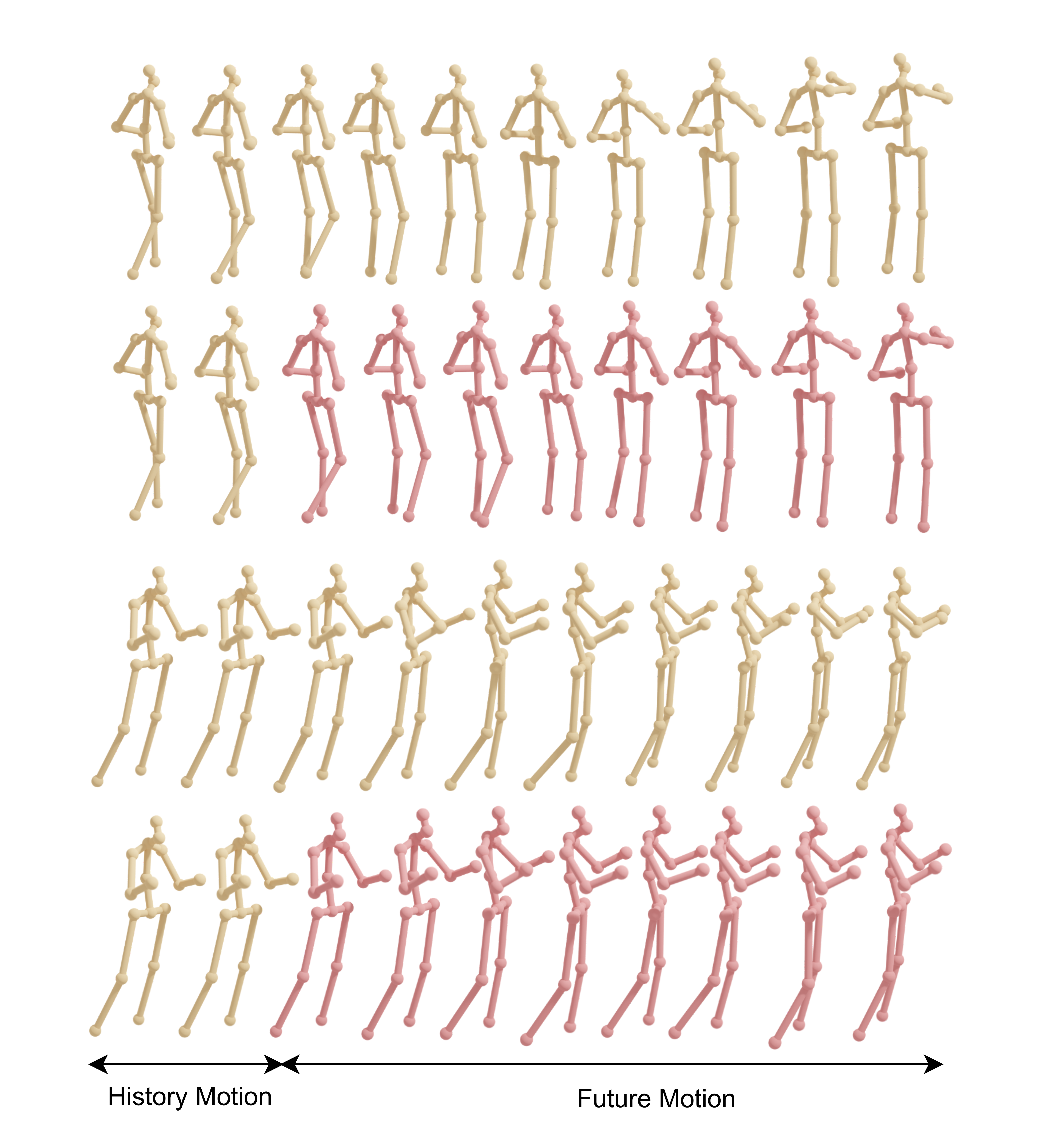} 
    \caption{Long-term predictions of \textit{MotionMixer} for the actions \textit{Directions} and \textit{Posing}  of the Human3.6M dataset. The first and the third line indicate the ground-truth 3D human motion. The frames on the left are the observations. The right part, shown in pink is the long-term motion prediction. One every three frames are shown. The model predictions accurately match the ground-truth body poses.}
    \label{fig:teaser}
\end{figure}

Recently, the availability of large-scale datasets, e.g.~Human3.6M \cite{ionescu2013human3}, AMASS \cite{mahmood2019amass} or 3DPW \cite{von2018recovering}, the development of human pose estimation algorithms ~\cite{belagiannis2014holistic,bouazizi2021self} and the advent of deep learning methods pushed the evolution towards forecasting future 3D poses with less priors. Several learning-based approaches were proposed to tackle the problem of 3D human motion prediction. Methods like ~\cite{fragkiadaki2015recurrent,martinez2017human,tang2018long} build upon the success of recurrent neural networks (RNNs) to better model the temporal correlation between the human body joints. Nevertheless, the use of RNNs come with many drawbacks, including vanishing or exploding gradients \cite{fragkiadaki2015recurrent}, first frame discontinuity, problems in processing longer input sequences as well as shorter forecast horizons leading to stationary predictions or frozen motion \cite{martinez2017human}. More recent works \cite{li2018convolutional,sofianos2021space} propose to replace RNN models with dedicated temporal convolutional architectures. In contrast to RNNs, Convolutional Neural Networks (CNNs) maintain a hierarchical structure, enabling them to capture both spatial and temporal correlations effectively. Lately, graph convolutional networks (GCNs) have received increasing attention. Several works \cite{mao2019learning,mao2020history,sofianos2021space,liu2021motion} attempted to utilize GCNs to learn fine-grained spatial relationships among joints.  \cite{mao2019learning} for instance, encoded the joints history in the frequency domain and proposed a GCN with learnable connectivity to predict the future motion. Although effective, these methods still need structural priors \cite{aksan2020spatio} or frequency transformation \cite{mao2019learning} to address the inherent spatial-temporal dependency of the human motion.

While RNNs, CNNs and GCNs led to a significant performance gain in forecasting 3D human body poses, the existing methods are unnecessarily complex. In this paper, we present \textit{MotionMixer}, the first model using exclusively MLPs to address the inherent problems of human motion. Equipped with a simple, yet effective architecture, the model aims to learn the spatial-temporal dependencies of the human body pose. Inspired by \cite{tolstikhin2021mlp}, we propose two types of layers: one with MLPs applied independently to time steps (i.e. “mixing” the temporal information) and another with MLPs applied across body poses (i.e. “mixing” the spatial information). The interchangeable spatial and temporal mixing operations allow the model to access current and past information directly and capture both the structural and the temporal dependencies explicitly.

Our contributions are summarized as follows:
1) We propose to jointly model the spatial locations of the body joints and their temporal dependency with a spatial-temporal MLP. To the best of our knowledge, \textit{MotionMixer} is the first 3D body pose forecasting approach based solely on MLPs. 2) We design an efficient architecture, that significantly reduces the computational cost of the pose forecasting model. 3) An extensive evaluation on three challenging large-scale datasets demonstrates state-of-the-art performance for short-term and long-term motion prediction.

\section{Related Work}

\paragraph{Recurrent-based Motion Prediction.}
3D human motion prediction with RNNs has been widely studied in the last years.  \cite{fragkiadaki2015recurrent} proposed a recurrent encoder-decoder model, which incorporates nonlinear encoder and decoder networks before and after recurrent layers. A curriculum learning strategy was adopted to prevent error accumulation during the training. To better model the long-term temporal dependency,  \cite{martinez2017human} incorporated a residual connection between the RNN units. To make reliable future predictions, \cite{tang2018long} proposed a motion context modeling by summarizing the historical human motion with respect to the current prediction within a recurrent prediction framework. Though these methods have also incorporated different modules into RNNs, exploring a recurrent-free backbone to address human motion prediction tasks is rarely studied.

\paragraph{Convolutional-based Motion Prediction.}
Thanks to their hierarchical structure and their effectiveness in capturing spatial and temporal correlations, there has been recently a great interest in integrating CNNs in human motion prediction. \cite{li2018convolutional} for instance, encoded the history motion into a long-term hidden variable, which is used with a decoder to predict the future sequence. The decoder itself also has an encoder-decoder structure, with a short-term encoder and a long-term decoder.  \cite{sofianos2021space} proposed a space-time-separable GCN, where the space-time graph connectivity is factored into space and time affinity matrices. \cite{mao2019learning} designed a fully connected GCN to adaptively learn the spatial connectivity of the human skeletons and converted the joint trajectory to the frequency domain to handle the temporal information. \cite{dang2021msr} proposed a multi-scale residual graph network with descending and ascending GCNs to extract features in both fine-to-coarse and coarse-to-fine manners. Despite the advantages in capturing long-range temporal correlations, the quite high computational cost of convolution-based approaches remains a bottleneck. Unlike these approaches, we propose an MLP-based model with lower computational complexity, that better exploits the spatial-temporal dependencies of the body pose. 

\paragraph{Attention- and MLP-based Architectures.}
Inspired by the success of the self-attention mechanism \cite{vaswani2017attention} in natural language processing, many works have explored its application in human motion prediction. \cite{mao2020history} proposed to capture the similarity between the current motion context and the historical motion sub-sequences in the frequency domain with the attention mechanism. \cite{aksan2020spatio} proposed to autoregressively learn spatial-temporal representations with  decoupled temporal and spatial self-attention. The key role of self-attention is to re-weight the relative importance of each pose in the sequence with respect to all other poses. This resulted in a high computational memory overhead with increasing number of history poses. On the other end of the spectrum, there have been new works that support replacing self-attention with MLPs. MLP-Mixer  \cite{tolstikhin2021mlp} for instance, relaxed the quadratically increasing computational memory by replacing the self-attention module with a two-layer MLP. The idea behind the Mixer architecture is to learn to separate the per-location operations and cross-location operations, allowing communication between different image patches. With a design based solely on MLPs, the model was originally developed for visual recognition tasks. However, unlike image classification, where only spatial correlations exist, there exist complicated spatial-temporal dynamics in human motion. In this work, we delve deeper and propose a new architecture based on MLPs to learn the spatial-temporal dependencies of the human body. Unlike the above-discussed approaches, we show that MLPs are effective in learning human dynamics. 

\section {Method}

We define the human body motion as a sequence of $T_h + T_f$ consecutive frames, where each frame parameterizes the angles or 3D coordinates of the human body joints. Let $\mathbf{X}_{1: T_h}=\left\{\mathbf{x}_{1}, \mathbf{x}_{2}, \ldots,\mathbf{x}_{t}, \ldots  \mathbf{x}_{T_h}\right\} \in \mathbb{R}^{3 \times J \times T_h}$ be the historical motion sequence until the current time step $T_h$, with the 3D body pose $\mathbf{x}_{t} \in \mathbb{R}^{3 \times J}$ and  $J$ is the number of body keypoints. Our goal is to learn the mapping that bridges the history sequence $\mathbf{X}_{1: T_h}$ to the future sequence  $\mathbf{X}_{T_h+1: T_h+T_f}=\left\{\mathbf{x}_{T_h+1}, \mathbf{x}_{T_h+2}, \ldots,\mathbf{x}_{T_h+t}, \ldots  \mathbf{x}_{T_h +T_f}\right\} \in \mathbb{R}^{3 \times J \times T_f}$ with a pure MLP-based network. Below, we provide the details of our network architecture.

\begin{figure*}[ht!]
    \centering
    \includegraphics[width=\linewidth]{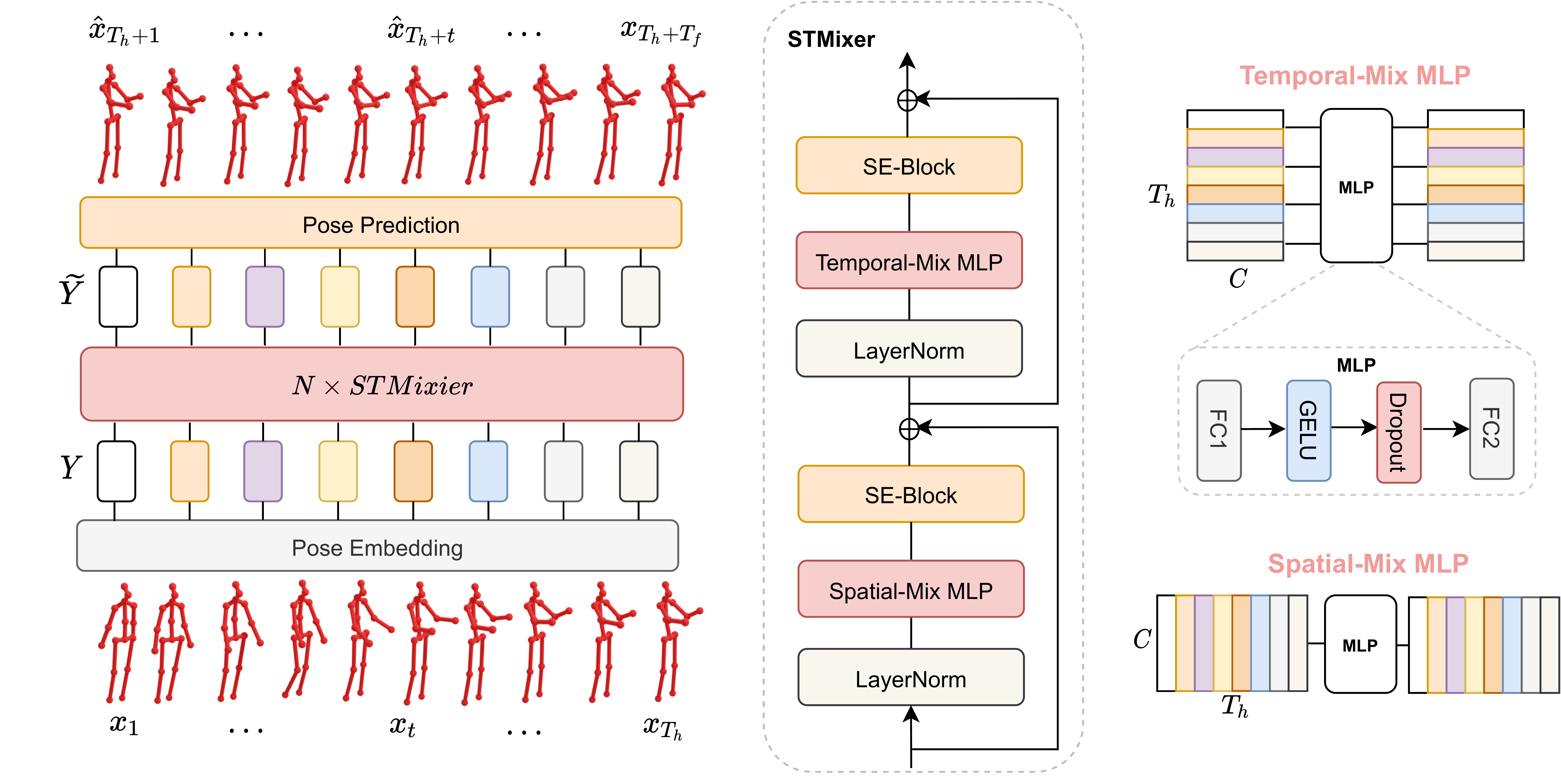} 
    \caption{\textbf{Overview of the proposed MotionMixer}. It mainly consists of three modules: pose embedding, spatial-temporal mixing, and pose prediction. First, the pose embedding module linearly projects each of the past 3D body poses through a single fully-connected layer to a hidden dimension $C$. The learned features are then fed to $N$ \textit{STMixer} layers. Equipped with a spatial-MLP, a temporal-MLP and a squeeze-and-excitation (SE) block, \textit{STMixer} aims to learn fine-grained spatial-temporal dependencies of the human motion. The mixing blocks are shown on the right. In each layer, we depict how our framework aggregates information via spatial-temporal mixing. An MLP-based body pose prediction is then applied to the mixed features to forecast the future human motion.}
    \label{fig:method}
\end{figure*}

\subsection{MotionMixer}
\textit{MotionMixer} is a sequence to sequence model with mainly three modules: pose embedding, spatial-temporal mixing, and pose prediction, as illustrated in Fig.~\ref{fig:method}. The pose embedding and the spatial-temporal mixing are coupled together to encode the spatial-temporal dependencies of the human body joints. The pose prediction module, which consists of two fully-connected layers decodes the future 3D motion. Given the historical sequence, each pose is first embedded by a fully-connected layer and given to repeated $N$ \textit{STMixer} blocks, each of which includes two MLPs with skip connections. The interaction of the body joints over time is modelled by two mixing operations within a single spatial-temporal MLP. The spatial-mixing allows the interplay between the spatial location of the joints, whereas the temporal-mixing allows the long-range interactions of the observed motion. In the pose prediction module, the outputs of the mixing are finally aggregated into a global vector and fed to an MLP to forecast the future motion. Below, we describe each module in detail.

\subsubsection{Pose Embedding} 

Given the observed motion sequence $\mathbf{X}_{1: T_h}$, the skeleton of each time step $\mathbf{x}_{t} $ is flattened into a vector of length $ K = 3 \times J$. This yields a two-dimensional tensor $\mathbf{X}_{1:T_h} \in \mathbb{R}^{T_h\times K}$ with one temporal dimension $T_h$ and one spatial dimension $K$. For simplicity, we omit the subscript $T_h$, thus replacing $\mathbf{X}_{1:T_h}$ with $\mathbf{X}$. The flattened sequence $\mathbf{X}$ is then processed by a learnable embedding, which linearly projects each body skeleton $\mathbf{x}_{t} \in \mathbb{R}^{K}$ through a single fully-connected layer to the hidden dimension $C$. We refer to the output of the learnable pose embedding $\mathbf{Y} = \left\{\mathbf{y}_{1},\ldots,\mathbf{y}_{t}, \ldots  \mathbf{y}_{T_h}\right\} \in \mathbb{R}^{C\times T_h} $ as: 
\begin{equation}
\mathbf{Y} = \mathbf{W}_0 \mathbf{X}  + \mathbf{b}_0, 
\end{equation}
where $\mathbf{W}_0 \in \mathbb{R}^{C\times T_h \times K}$ and $\mathbf{b}_0 \in \mathbb{R}^{C \times T_h}$ are weights of the fully-connected layer. 

\subsubsection{Spatial-Temporal Mixing}
 The proposed mixing module is motivated by the fact that the human spatial-temporal dynamics are contiguous. The spatial-temporal mixing stacks $N$ \textit{STMixer} blocks of identical size and structure. Each block, as shown in Fig.~\ref{fig:method} includes two types of MLP layers: spatial-MLP and temporal-MLP, each followed by a squeeze-and-excitation block \cite{hu2018squeeze}. The spatial-mixing aims to learn fine-grained spatial dependencies between the body joints by acting on the columns of the pose embedding $\mathbf Y$. Each column encodes the spatial information of one timestep. The spatial-mixing operation can be written as follows:
\begin{equation}
\hat{\mathbf{Y}} = \mathbf{Y} + \mathbf{W}_2 \sigma(\mathbf{W}_1\text{LN}(\mathbf{Y})),
\end{equation}
 where $\mathbf{W}_1 \in \mathbb{R}^{C \times C}$, $\mathbf{W}_2 \in \mathbb{R}^{{C} \times C}$, $\sigma (\cdot)$ is a $\text{GELU}$ activation function \cite{hendrycks2016gaussian} and $\text{LN} (\cdot)$ denotes the layer normalization \cite{ba2016layer}.
 Driven by the fact that the human body joints contribute unequally to the forecasted motion, mixing the columns of $\mathbf Y$ allows the communication between different spatial pose embeddings. To enable the interchanging between the spatial and temporal domains, the spatial-mixed skeleton features $\hat{\mathbf{Y}}$ are transposed and fed to the temporal-mixing MLP. The temporal-mixing MLP, by acting on rows of $\hat{\mathbf{Y}}$ aims to learn the temporal correlation of the spatial-mixed features. The temporal-mixing operation is shared across all rows, which encode the temporal information of one body joint. Formally, this can be written as follows: 
  \begin{equation}
\widetilde{\mathbf{Y}} = \hat{\mathbf{Y}} + (\mathbf{W}_4 \sigma(\mathbf{W}_3\text{LN}(\hat{\mathbf{Y}}^\top)))^\top,
\end{equation}
 where $\mathbf{W}_3 \in \mathbb{R}^{T_h\times T_h}$, $\mathbf{W}_4 \in \mathbb{R}^{T_h\times T_h}$. Each linear operator of the temporal-mixing assigns each time step as a linear combination of all frames where the linear weights depend on the frame’s location.  As such, the temporal information can be maintained in each mixing step, allowing the model to capture long-term dependencies by applying long-range interactions between frames.

In the motion prediction, each time step has a different contribution that is not known in advance. We introduce a squeeze-and-excitation (SE) block \cite{hu2018squeeze} into \textit{STMixer} to automatically regulate the input importance. The SE block, as shown in Fig.~\ref{fig:method} is added after each mixing operation helping the network to re-weight the influence of each time step. Formally, this is defined as:
\begin{align}
\label{eq:seq_exc}
\hat{\mathbf{Y}} = \mathbf{Y} + \delta (\mathbf{W}_{s} \sigma_R (\mathbf{W}_{e} (\mathbf{W}_2 \sigma(\mathbf{W}_1\text{LN}(\mathbf{Y})))),\\
\widetilde{\mathbf{Y}} = \hat{\mathbf{Y}} + \delta (\mathbf{W}_{s} \sigma_R (\mathbf{W}_{e} (\mathbf{W}_4 \sigma(\mathbf{W}_3\text{LN}(\hat{\mathbf{Y}}^\top)))^\top),\nonumber
\end{align}
where $\delta (\cdot) $ and $ \sigma_R (\cdot)$ are respectively the $\text{Softmax}$  and $\text{ReLU}$ activation functions. The weights $\mathbf{W_s} \in \mathbb{R}^{s\times e}$ and $\mathbf{W_e} \in \mathbb{R}^{e\times s}$ are shared across the spatial and temporal mixing units. After the mixing operation, an MLP-based pose prediction learns to generate the future human motion. 

\subsubsection{Pose Prediction}
Let $\mathbf{\widetilde{Y}}$ be the output features of the spatial-temporal mixing. An MLP-based decoder further propagates the mixed features for future pose forecasting. Each $\mathbf{\widetilde{y}}_t$ of $\mathbf{\widetilde{Y}}$ is projected to a vector of length $T_f$ based on a two-layer non-linear feed-forward network. The computation of the predicted body poses $\mathbf{\hat X}_{T_h+1: T_h+T_f}$ is described as:
\begin{equation}
\mathbf{\hat X}_{T_h+1: T_h+T_f} = \mathbf{W}_{p2}(\sigma_{R}(\mathbf{W}_{p1}(\mathbf{\widetilde{Y}})+\mathbf{b}_{p1}))+ \mathbf{b}_{p2},
\end{equation}
where $\mathbf{W}_{p1} \in \mathbb{R}^{C\times T_f}$, $\mathbf{W}_{p2} \in \mathbb{R}^{3\times J\times T_f}$ and $\mathbf{b}_{p2} \in \mathbb{R}^{3\times J}$, $\mathbf{b}_{p2} \in \mathbb{R}^{C}$ are the weights of the fully-connected layers. 

\subsection{Training}

 Each joint is represented by the position displacement between two adjacent frames. We train our model by predicting future displacements, which are then added to the most recent pose to get the full-body pose sequence. The loss in terms of Mean Per Joint Position Error (MPJPE) is given by:
\begin{equation}\label{input_loss}
\mathcal{L}_{3D} = \frac{1}{J \times T_f}  \sum_{j=1}^{J} \sum_{t=T_h +1 }^{T_h+T_f}  \parallel \Delta \mathbf{\hat{x}}_{t,j}  - \Delta \mathbf{x}_{t,j} \parallel_2,
\end{equation}
with $\Delta \mathbf{\hat{x}}_{t,j}$ and $\Delta \mathbf{x}_{t,j}$ denoting the predicted and ground-truth displacement of a joint $j$ between two adjacent frames. $||\cdot||_2$ indicates the $\ell_2$ norm. For the angle-based representation, the loss between the predicted joint angles and the ground truth in the exponential map representation   is given by:
\begin{equation}\label{input_mae}
\mathcal{L}_{MAE} = \frac{1}{J \times T_f}  \sum_{j=1}^{J} \sum_{t=T_h +1}^{T_h+T_f}  \parallel \mathbf{\hat{x}}_{t,j}  -  \mathbf{x}_{t,j} \parallel_2,
\end{equation}
where  $ \mathbf{\hat{x}}_{t,j}$ denotes the predicted angle of the joint $j$ at frame $t$ and $ \mathbf{x}_{t,j}$ the corresponding ground-truth.  

\section {Experimental Evaluation}
We evaluate our model on three large-scale public benchmarks. Below, we first introduce the datasets, the evaluation metrics and the baselines we compare with. We then present our results using 3D coordinates and joint angles.

\subsection{Datasets and Metrics}
\paragraph{Human3.6M.} \cite{ionescu2013human3} consists of 7 actors performing 15 different actions. The original data is transformed from exponential map to 3D joint coordinates. We consider 22 joints for forecasting the 3D body poses and 16 for the angle-based prediction.  Following \cite{sofianos2021space,mao2020history}, we use the subject (S11) for validation, (S5) for testing, and the rest of the subjects for training.
\paragraph{AMASS.} \cite{mahmood2019amass} is a recently published dataset, which consists of 40 subjects performing the action of \textit{walking}. Following \cite{sofianos2021space,mao2021multi}, we select 8 datasets for training, 4 for validation, and one (BMLrub) as the test set. For each body pose, we consider 18 joints.
\paragraph{3DPW.} The 3D Pose in the Wild dataset \cite{von2018recovering} consists of video sequences acquired by a moving phone camera. Overall, it contains 51,000 frames of indoor and outdoor actions captured at 30Hz. We use the official test set to test the generalization of a model trained on AMASS.

\begin{table*} [ht]
\centering
\resizebox{0.99\textwidth}{!}{%
\begin{tabular}{|c|ccccc|ccccc|ccccc|ccccc|}
  & \multicolumn{5}{c|}{Walking} & \multicolumn{5}{c|}{Eating} & \multicolumn{5}{c|}{Smoking} & \multicolumn{5}{c|}{Discussion} \\
     milliseconds       & 80   & 160  & 320  & 400 &1000 & 80   & 160  & 320  & 400  &1000 & 80   & 160  & 320  & 400  & 1000  & 80   & 160  & 320  & 400 &1000    \\\hline
Res. sup ~\cite{martinez2017human} &  23.2 & 40.9 & 61.0 & 66.1 & 79.1 & 16.8 & 31.5 & 53.5 & 61.7 & 98.0  & 18.9 & 34.7 & 57.5 & 65.4 & 102.1  & 25.7 & 47.8 & 80.0 & 91.3 & 131.8 \\
convSeq2Seq ~\cite{li2018convolutional} & 17.7 & 33.5 & 56.3 & 63.6 &  82.3 & 11.0 & 22.4 & 40.7 & 48.4 &  87.1 & 11.6 & 22.8 & 41.3 & 48.9 & 81.7  & 17.1 & 34.5 & 64.8 & 77.6  & 129.3 \\
LTD-10-25 \cite{mao2019learning} & 12.3 & 23.2 & 39.4 & 44.4  & 60.9 & 7.8 & 16.3 & 31.3 & 38.6  & 75.8 & 8.2 & 16.8 & 32.8 & 39.5  & 72.1 & 11.9 & 25.9 & 55.1 & 68.1  & 118.5 \\
RNN-GCN \cite{mao2020history}  & 10.0 & 19.5 & 34.2 & 39.8  & 58.1 & 6.4 & 14.0 & 28.7 & 36.2  & 75.7 & 7.0 & 14.9 & 29.9 & 36.4  &  69.5 & 10.2 & 23.4 & 52.1 & 65.4  & 119.8 \\
MSRGCN \cite{dang2021msr} & 12.1 & 22.6 & 38.6 & 45.2  & 63.0  & 8.3 & 17.0 & 33.0 & 40.4  & 77.1 & 8.0 & 16.2 & 31.3 & 38.1  & 71.6   & 11.9 & 26.7 & 57.0 & 69.7  & 117.6 \\
MultiAttention \cite{mao2021multi} & 9.9   & 19.3    & 33.7   & 39.0  & 57.1  & 7.9 & 17.5 & 37.4 & 45.2  & 73.7 & 7.0 & 14.3 & 25.4  & 29.0  & 68.7   & 8.6 & 22.8 & 51.0 & 64.0 & 117.5 \\
STSGCN \cite{sofianos2021space} $\dagger$  & 10.7 & 16.8 & 29.1 & 38.2  & 51.8 & 6.7 & 11.3 & 22.6 & 31.6  & 52.5 & 7.1 & 11.6 & 22.3 & 30.6  & 50.1 & 9.7 & 16.7 & 33.4 & 45.0  & 78.8\\
GAGCN \cite{zhong2022spatial} $\dagger$ & 10.3 & 16.1 & 28.8 & 32.4  & 51.1  & 6.4 & 11.5 & 21.7 & 25.2  & 51.4 & 7.1  & 11.8  & 21.7 & 24.3   & 48.7   & 9.7  & 17.1  & 31.4  & 38.9   & 76.9 \\\hline
Ours & 10.8 & 22.4 & 36.5 & 42.4 & 59.9 &  7.7 & 14.0 & 27.3 & 36.1  & 76.6 & 7.1 & 14.0 & 29.1 & 36.8  & 68.5 & 10.2 & 22.5 & 51.0 & 64.1   & 117.4
\\
Ours $\dagger$  & \textbf{7.3} & \textbf{12.9} & \textbf{23.5} & \textbf{28.6} & \textbf{49.2} &  \textbf{4.3} & \textbf{8.3} & \textbf{16.9} & \textbf{20.9}  & \textbf{47.4} & \textbf{4.7} & \textbf{8.8} & \textbf{17.3} & \textbf{21.4}  & \textbf{45.4} & \textbf{6.4} & \textbf{13.1} & \textbf{28.6} & \textbf{35.5}   & \textbf{78.0}
\\\hline
 & \multicolumn{5}{c|}{Directions}  & \multicolumn{5}{c|}{Greeting} & \multicolumn{5}{c|}{Phoning} & \multicolumn{5}{c|}{Posing} \\
     milliseconds       & 80   & 160  & 320  & 400 &1000 & 80   & 160  & 320  & 400  &1000 & 80   & 160  & 320  & 400  & 1000  & 80   & 160  & 320  & 400 &1000   \\\hline
Res. sup ~\cite{martinez2017human} &  21.6 & 41.3 & 72.1 & 84.1 & 129.1 & 31.2 & 58.4 & 96.3 & 108.8 &  153.9 & 21.1 & 38.9 & 66.0 & 76.4 &  126.4 & 29.3 & 56.1 & 98.3 & 114.3 & 183.2\\
convSeq2Seq ~\cite{li2018convolutional} & 13.5 & 29.0 & 57.6 & 69.7 & 115.8 & 22.0 & 45.0 & 82.0 & 96.0 & 147.3 & 13.5 & 26.6 & 49.9 & 59.9 & 114.0 & 16.9 & 36.7 & 75.7 & 92.9  & 187.4  \\
LTD-10-25 \cite{mao2019learning} & 8.8 & 20.3 & 46.5 & 58.0  & 105.5 & 16.2 & 34.2 & 68.7 & 82.6  & 136.8 & 9.8 & 19.9 & 40.8 & 50.8  & 105.1 & 12.2 & 27.5 & 63.1 & 79.9  & 174.8 \\
RNN-GCN \cite{mao2020history}   & 7.4 & 18.4 & 44.5 & 56.5   & 106.5 & 13.7 & 30.1 & 63.8 & 78.1  & 138.8 & 8.6 & 18.3 & 39.0 & 49.2  & 105.0 & 10.2 & 24.2 & 58.5 & 75.8  & 178.2 \\
MSRGCN \cite{dang2021msr} & 8.6 & 19.6 & 43.2 & 53.8  & 100.6  & 16.4 & 36.9 & 77.3 & 93.3  & - & 10.1 & 20.7 & 41.5 & 51.2  & -   & 12.8 & 29.4 & 66.9 & 85.0  & - \\
MultiAttention \cite{mao2021multi} & 11.3 & 22.9 & 50.6 & 62.6  & 105.7 & 12.9 & 26.6 & 68.2 & 85.4 & 136.7 & 11.2 & 19.6 & 37.7 & 44.1   & 104.6 & 9.8  & 23.7 & 62.2 & 78.7 & 172.9 \\
STSGCN \cite{sofianos2021space} $\dagger$ & 7.4 & 13.5 & 29.2 & 40.9  & 71.0 & 12.4 & 21.7 & 42.1 & 54.5  & 91.6 & 8.2 & 13.7 & 26.8 & 36.6  & 66.1 & 9.9 & 18.0 & 38.2 & 52.6   & 106.4\\
GAGCN \cite{zhong2022spatial} $\dagger$ & 7.3  & 12.8  & 30.3  & 34.5   &  69.9  & 11.8  & 20.1  & 40.5  & 48.4   & 87.7 & 8.8  & 13.5  & 25.5  & 28.7   &  66.0   & 10.1   & 17.0  & 35.5  & 45.1 &  99.1\\\hline
Ours & 8.3 & 18.1 & 43.8 & 53.4 & 105.4 &  12.8 & 33.4 & 62.3 & 82.2  & 136.5 & 10.0 & 20.1 & 37.4 & 51.1  & 104.4 & 11.7 & 23.3 & 62.4 & 79.5   & 174.9
\\
Ours  $\dagger$ & \textbf{4.4} & \textbf{9.7} & \textbf{22.5} & \textbf{29.2} & \textbf{66.5} &  \textbf{8.8} & \textbf{17.7} & \textbf{36.9} & \textbf{46.2}  & \textbf{93.6} & \textbf{5.6} & \textbf{10.7} & \textbf{21.9} & \textbf{27.8}  & \textbf{63.4} & \textbf{6.0} & \textbf{13.1} & \textbf{30.2} & \textbf{40.1}    & \textbf{99.7}
\\\hline
 & \multicolumn{5}{c|}{Purchases} & \multicolumn{5}{c|}{Sitting} & \multicolumn{5}{c|}{Sitting Down} & \multicolumn{5}{c|}{Taking Photo} \\
     milliseconds       & 80   & 160  & 320  & 400 &1000 & 80   & 160  & 320  & 400  &1000 & 80   & 160  & 320  & 400  & 1000  & 80   & 160  & 320  & 400 &1000   \\\hline
Res. sup ~\cite{martinez2017human}&  28.7 & 52.4 & 86.9 & 100.7 & 154.0 & 23.8 & 44.7 & 78.0 & 91.2 & 152.6 & 31.7 & 58.3 & 96.7 & 112.0 & 187.4 & 21.9 & 41.4 & 74.0 & 87.6 & 153.9\\
convSeq2Seq ~\cite{li2018convolutional} & 20.3 & 41.8 & 76.5 & 89.9 & 151.5 & 13.5 & 27.0 & 52.0 & 63.1 & 120.7 & 20.7 & 40.6 & 70.4 & 82.7 & 150.3 & 12.7 & 26.0 & 52.1 & 63.6  & 128.1  \\
LTD-10-25 \cite{mao2019learning} & 15.2 & 32.9 & 64.9 & 78.1  & 134.9 & 10.4 & 21.9 & 46.6 & 58.3  & 118.7 & 17.1 & 34.2 & 63.6 & 76.4  & 143.8 & 9.6 & 20.3 & 43.3 & 54.3  & 115.9 \\
RNN-GCN \cite{mao2020history}   & 13.0 & 29.2 & 60.4 & 73.9  & 135.9 & 9.3 & 20.1& 44.3 & 56.0  & 138.8 & 14.9 & 30.7 & 59.1 & 72.0  & 143.6 & 8.3 & 18.4 & 40.7 & 51.5  & 115.9 \\
MSRGCN  \cite{dang2021msr} & 14.7 & 32.4 & 66.1 & 79.6  & -  & 10.5 & 21.9 & 46.2 & 57.8 & - & 16.1 & 31.6 & 62.4 & 76.8  & -   & 9.8 & 21.0 & 44.5 & 56.3 & - \\
MultiAttention \cite{mao2021multi} & 18.1 & 36.8 & 58.4 & 67.9  &  133.1 & 9.9 & 24.3 & 53.8 & 66.3 & 115.0 & 10.4 & 26.6 & 54.6  & 66.3 & 141.8 & 5.9 & 14.8 & 38.0 & 49.4 & 115.2 \\
STSGCN  \cite{sofianos2021space} $\dagger$ & 11.9 & 21.3 & 41.9 & 54.8  & 93.5 & 9.1 & 15.1 & 29.8 & 39.8  & 75.3 & 14.4 & 23.7 & 41.9 & 53.8  & 94.3 & 8.1 & 14.1 & 29.7 & 41.9  & 76.9 \\
GAGCN \cite{zhong2022spatial} $\dagger$ & 11.9  & 20.7  & 41.8  & 47.6   & 85.1  & 9.3   & 14.4   & 29.6   & 38.5   & 71.1 & 14.1   & 24.8 & 40.0   & 47.4  & 84.1   & 8.5  & 13.9  & 28.8  & 35.1    & 70.0 \\\hline
Ours  & 14.6 & 31.3 & 62.8 & 76.1 & 135.1 &  10.0 & 20.9 & 43.7 & 54.5  & 115.7 & 12.0 & 31.4 & 61.4 & 74.5  & 141.1 & 9.0 & 18.9 & 41.0 & 51.6   & 114.6
\\
Ours  $\dagger$ & \textbf{8.4} & \textbf{16.9} & \textbf{34.1} & \textbf{42.7} & \textbf{88.7} &  \textbf{6.5} & \textbf{11.8} & \textbf{23.6} & \textbf{29.8}  & \textbf{68.9} & \textbf{10.9} & \textbf{18.8} & \textbf{35.1} & \textbf{42.6}  & \textbf{89.3} & \textbf{5.5} & \textbf{10.4} & \textbf{22.1} & \textbf{27.9}   & \textbf{66.6}
\\\hline
 & \multicolumn{5}{c|}{Waiting} & \multicolumn{5}{c|}{Walking Dog} & \multicolumn{5}{c|}{Walking Together} & \multicolumn{5}{c|}{Average} \\
     milliseconds       & 80   & 160  & 320  & 400 &1000 & 80   & 160  & 320  & 400  &1000 & 80   & 160  & 320  & 400  & 1000  & 80   & 160  & 320  & 400 &1000   \\\hline
Res. sup  ~\cite{martinez2017human} &  23.8 & 44.2 & 75.8 & 87.7 & 135.4 & 36.4 & 64.8 & 99.1 & 110.6 & 164.5 & 20.4 & 37.1 & 59.4 & 67.3 & 98.2 & 25.0 & 46.2 & 77.0 & 88.3 & 136.6\\
convSeq2Seq  ~\cite{li2018convolutional} & 14.6 & 29.7 & 58.1 & 69.7 & 117.7 & 27.7 & 53.6 & 90.7 & 103.3 & 162.4 & 15.3 & 30.4 & 53.1 & 61.2 & 87.4 & 16.6 & 33.3 & 61.4 & 72.7 & 124.2  \\
LTD-10-25  \cite{mao2019learning} & 12.3 & 23.2 & 39.4 & 44.4  & 108.3 & 7.8 & 16.3 & 31.3 & 38.6  & 146.4 & 8.2 & 16.8 & 32.8 & 39.5  & 65.7 & 11.9 & 25.9 & 55.1 & 68.1  & 112.4 \\
RNN-GCN \cite{mao2020history}  & 8.7 & 19.2 & 43.4 & 54.9  & 108.2 & 20.1 & 40.3 & 73.3 & 86.3  & 146.9 & 8.9 & 18.4& 35.1 & 41.9 & 64.9 & 10.4 & 22.6 & 47.1 & 58.3  & 112.1 \\
MSRGCN \cite{dang2021msr} & 10.6 & 23.0 & 48.2 & 59.2  & -  & 20.6 & 42.8 & 80.3 & 93.3  & - & 10.5 & 20.9 & 37.4 & 43.8  & 65.9   & 12.1 & 25.5 & 51.6 & 62.9  & 114.2 \\
MultiAttention \cite{mao2021multi} & 9.0 & 22.5 & 55.7 & 71.1 & 105.1 & 29.5 & 54.8 & 100.3 & 119.0  & 141.4 & 8.0 & 17.6 & 33.2 & 42.0 & 63.2  & 11.0 & 23.6 & 49.2 & 60.0  & 110.1 \\
STSGCN  \cite{sofianos2021space} $\dagger$ & 8.6 & 14.7 & 29.6 & 40.7  & 72.0 &17.6 & 29.3 & 52.6 & 66.4  & 102.6 & 8.6 & 14.3 & 26.5 & 35.1 & 51.1 & 10.1 & 17.1 & 33.1 & 38.3 & 75.6\\
GAGCN \cite{zhong2022spatial} $\dagger$ & 8.5  & 14.1  & 29.8  & 33.8   & 69.3   & 17.0  & 28.8  & 50.1  & 59.4   & 91.3 & - & - & - & -  & -   & 10.1  & 16.9  & 32.5  & 38.5   & 72.9\\\hline
Ours  & 10.2 & 21.1 & 45.2 & 56.4 & 107.7 &  20.5 & 42.8 & 75.6 & 87.8 & 142.2 & 10.5 & 20.6 & 38.7 & 43.5  & 65.4 & 11.0 & 23.6 & 47.8 & 59.3   & 111.0
\\

Ours  $\dagger$ & \textbf{5.4} & \textbf{10.9} & \textbf{23.2} & \textbf{30.0} & \textbf{68.2} &  \textbf{13.4} & \textbf{24.6} & \textbf{45.2} & \textbf{54.1} & \textbf{99.6} & \textbf{5.9} & \textbf{11.3} & \textbf{22.2} & \textbf{27.4}  & \textbf{50.4} & \textbf{9.0} & \textbf{13.2} & \textbf{26.9} & \textbf{33.6}   & \textbf{71.6}
\\\hline

\end{tabular}
}
\caption{Performance comparison between different methods in terms of short-term and long-term pose prediction via mean per joint position error for each activity from the Human3.6M dataset. We provide the error results for the particular frame as well as the average over all frames. ($\dagger$) indicates methods that compute the average error over all frames. All other approaches evaluate at the particular frame, where the error is measured between the predictions and ground truth at each frame. The best performance is highlighted in boldface.} 
\label{tab:h36_all_3d}
\end{table*}

\paragraph{Metrics.} Following the standard evaluation protocol \cite{li2018convolutional,mao2020history,sofianos2021space}, we report the euclidean distance between the predicted and ground-truth joint angles. Due to the inherent ambiguity of the Euler-angle representation \cite{mao2019learning,mao2020history}, we further report results in terms of 3D error. We make use of the Mean Per Joint Position Error (MPJPE) in millimeters. We provide the results at the particular frame, as well as the average over all frames following \cite{sofianos2021space,zhong2022spatial}. For the particular frame evaluation, the MPJPE is measured between the predicted pose sequence and the corresponding ground truth at each frame, whereas for the average frame evaluation, the errors in all previous frames w.r.t. a considered one are computed and then averaged.

\subsection{Implementation Details}
\textit{MotionMixer} contains three \textit{STMixer} blocks with  $C = 60$ channels. In each MLP block, a dropout layer with a rate of 0.1 is added to prevent overfitting. We use Adam \cite{kingma2014adam} as the optimizer. During training, the learning rate is set to $10^{-2}$ and decayed by a factor of 0.1 every 10 epochs.  We train our model for 50 epochs with a batch size of 50 for Human3.6M and 256 for AMASS. 

\subsection{Baselines } 
Following previous works \cite{martinez2017human,li2018convolutional,mao2019learning}, we quantitatively evaluate our proposed model on all kinds of actions against the state-of-the-art for 400ms short-term  ({\em i.e.}, 10 frames) and 1000ms long-term  ({\em i.e.}, 25 frames) predictions. 
We include nine methods with recurrent \cite{martinez2017human,mao2020history}, convolutional \cite{li2018convolutional,tang2018long}, graph-convolutional \cite{mao2019learning,sofianos2021space,dang2021msr,zhong2022spatial}, and attention-based architectures  \cite{mao2021multi} in our comparison . 

\subsection{Results}

\begin{table*}[htp]
\centering
\resizebox{0.9\textwidth}{!}{%
\begin{tabular}{|c|cccccccc||cccccccc|}
& \multicolumn{8}{c||}{Average  3D} & \multicolumn{8}{c|}{Average  MAE} \\
milliseconds   & 80   & 160  & 320  & 400  & 560 & 720 & 880 & 1000 & 80   & 160  & 320  & 400  & 560 & 720 & 880 & 1000 \\\hline
Res. sup.~\cite{martinez2017human} & 25.0 & 46.2 & 61.4 & 88.3 &  106.3 & 119.4 & 130.0 & 136.6 & 0.36 & 0.67 & 1.02 & 1.15 &  - & - & - & - \\
convSeq2Seq~\cite{li2018convolutional} & 16.6 & 33.3 & 77.0 & 72.7 &  90.7 & 104.7  & 116.7  & 124.2 & 0.38 & 0.68 & 1.01 & 1.13 & 1.35 & 1.50 & 1.69 & 1.82 \\
MHU~\cite{tang2018long} & - & - & - & - &  - & -  & -  & - & 0.39 & 0.68 & 1.01 & 1.13 & 1.14 & 1.28 & 1.46 & 1.57 \\
LTD-10-25~\cite{mao2019learning} & 11.2 & 23.4 & 47.9 & 58.9 & 78.3  & 93.3  & 106.0  & 114.0 & 0.32 & 0.55 & 0.91 & 1.04 & 1.26 & 1.44 & 1.59 & 1.68\\
RNN-GCN \cite{mao2020history}  & 10.4 & 22.6 & 47.1 & 58.3 & 77.3 & 91.8  & 104.1 & 112.1 & 0.31 & 0.55 & 0.90 & 1.04 & 1.25 & 1.42  & 1.56 & 1.65 \\
Motion-Attention \cite{mao2021multi}  & 11.0 & 23.6 & 49.1 & 60.0 & 75.9 & 90.4 & 102.5 & 110.1 &  0.27  & 0.51 & 0.81 & 0.93 &  1.12 & 1.27 & 1.46 & 1.57\\
STSGCN \cite{sofianos2021space}($\dagger$)  & 10.1 & 17.1 & 33.1 & 38.3 & 50.8 & 60.1 & 68.9 & 75.6 &  0.24 & 0.39 & 0.59 & 0.66 & 0.79 & 0.92 & 1.00 & 1.09\\
GAGCN \cite{zhong2022spatial}($\dagger$)  & 10.1 & 16.9 & 32.5 & 38.5 & 50.0 & - & - & 72.9 &  0.24  & 0.38  & \textbf{0.54}  & 0.65  & \textbf{0.74} & - & - & \textbf{1.02} \\\hline

Ours   & 11.0 &  23.6  &  47.8  &  59.3  &  77.8  &  91.4  &  106.0  &  111.0  &    0.29  &  0.54  &  0.81  &  0.94  &  1.20  &  1.30  &  1.40  &  1.57  \\
Ours  ($\dagger$)  & \textbf{6.9} & \textbf{13.2} & \textbf{26.9} & \textbf{33.6} & \textbf{46.1} & \textbf{56.5} & \textbf{65.7} & \textbf{71.6}  &   \textbf{0.20} & \textbf{0.34} & 0.55 & \textbf{0.63} & 0.78 & \textbf{0.91} & \textbf{0.99} & 1.08 \\\hline
\end{tabular}
}
\caption{Average short-term and long-term 3D and mean angle prediction errors over all actions of Human3.6M. We provide the error results for the particular frame as well as the average over all frames. ($\dagger$) indicates methods that compute the average error over all frames. All other approaches evaluate at the particular frame, where the error is measured between the predictions and ground truth at each frame. The best performance is highlighted in boldface.}
\label{tab:mae_3d}
\end{table*}

\begin{table*}[htp]
\centering
\resizebox{0.9\textwidth}{!}{%
\begin{tabular}{|c|cccccccc||cccccccc|}
& \multicolumn{8}{c||}{AMASS-BMLrub} & \multicolumn{8}{c|}{3DPW} \\
milliseconds   & 80   & 160  & 320  & 400  & 560 & 720 & 880 & 1000 & 80   & 160  & 320  & 400  & 560 & 720 & 880 & 1000 \\\hline
convSeq2Seq~\cite{li2018convolutional} & 20.6 & 36.9 & 59.7 & 67.6 & 79.0 & 87.0 & 91.5 & 93.5 & 18.8 & 32.9 & 52.0 & 58.8 & 69.4 & 77.0 & 83.6 & 87.8 \\
LTD-10-25~\cite{mao2019learning} & 11.0 & 20.7 & 37.8 & 45.3 & 57.2 & 65.7 & 71.3 & 75.2 & 12.6 & 23.2 & 39.7 & 46.6 & 57.9 & 65.8 & 71.5 & 75.5 \\
RNN-GCN \cite{mao2020history}  & 11.3 & 20.7 & 35.7 & 42.0 & 51.7 & 58.6 & 63.4 & 67.2 & 12.6 & 23.1 & 39.0 & 45.4 & 56.0 & 63.6 & 69.7 & 73.7 \\
Motion-Attention \cite{mao2021multi}  & 11.0 & 20.3 & 35.0 & 41.2 & 50.7 & 57.4 & 61.9 & 65.8 &  12.4 & 22.6  & 38.1 & 44.4 & 54.7 & 62.1 & 67.9 & 71.8\\
STSGCN \cite{sofianos2021space}  & 10.0 & 12.5 & 21.8 & 24.5 & 31.9 & 38.1 & 42.7 & 45.5 &  8.6 & 12.8 & 21.0 & 24.5 & 30.4 & 35.7 & 39.6 & 42.3 \\
\hline
Ours  &  \textbf{6.6} & \textbf{10.3} & \textbf{18.0} & \textbf{21.9} & \textbf{28.8} & \textbf{33.6} & \textbf{38.8} & \textbf{41.6} &  \textbf{7.4} & \textbf{11.4} & \textbf{19.3} & \textbf{22.8} & \textbf{29.3} & \textbf{34.6} & \textbf{39.0} & \textbf{42.1} \\\hline
\end{tabular}
}
\caption{Short-term and long-term prediction of 3D body poses on AMASS-BMLrub (left) and 3DPW (right). All results are in millimeters. The best performance is highlighted in boldface.}
\label{tab:amass_3d}
\end{table*}

\paragraph{Human3.6M.} In Tab.~\ref{tab:h36_all_3d}, we provide the results for each activity of the Human3.6M dataset using 3D body poses. \textit{MotionMixer} outperforms all previous methods for short-term and long-term prediction in the average-frame evaluation. In particular, we outperform the current best-performing approach \cite{zhong2022spatial} by at least $1.3\%$ over all time horizons. The gain over the second-best approach \cite{sofianos2021space} ranges from $13\%$ in the case of 400ms, up to $5\%$ for 1000ms. In the particular frame evaluation, we reach very competitive results to the state-of-the-art approaches. Only \cite{mao2021multi} outperforms \textit{MotionMixer} in long-term prediction. Our method, however, yields larger improvements on activities with more complex dynamics such as \textit{WalkingDog} or \textit{Purchase}. Also on highly aperiodic actions like \textit{Posing}, our model still produces accurate predictions. Fig.~\ref{fig:teaser} illustrates the future predictions of the \textit{Posing} action. The predicted skeletons accurately match the ground-truth body poses. This demonstrates the effectiveness of  spatial-temporal mixing in learning fine-grained motion patterns.  In Tab.~\ref{tab:mae_3d}, we additionally provide the results over all actions using respectively the 3D body poses and the joint angles. Despite the inherent ambiguity of the angle-based representation, our method outperforms the compared methods in short-term and yields the lowest angle error of 0.2 at 80ms. In long-term prediction, we reach comparable results with previous approaches.

\paragraph{AMASS \& 3DPW.} The results of short-term and long-term prediction in 3D on AMASS and 3DPW are shown in Tab.~\ref{tab:amass_3d}. \textit{MotionMixer} gets the best average error at all short-term forecast times on the AMASS dataset. For long-term prediction, our method consistently outperforms all previous approaches. The performance gain ranges from $4\%$ for 1000ms up to $36\%$ for 80ms, which further shows the benefits of the proposed spatial-temporal mixing. We further test the generalization of a model trained on AMASS on 3DPW. Without any fine-tuning, our approach outperforms previous approaches at different forecast times and can therefore better generalize to complex outdoor environments.

\subsection{Ablation Studies}

\begin{figure} [ht] 
    \centering
    \vspace{-0.1in}
    \includegraphics[width=0.48\textwidth]{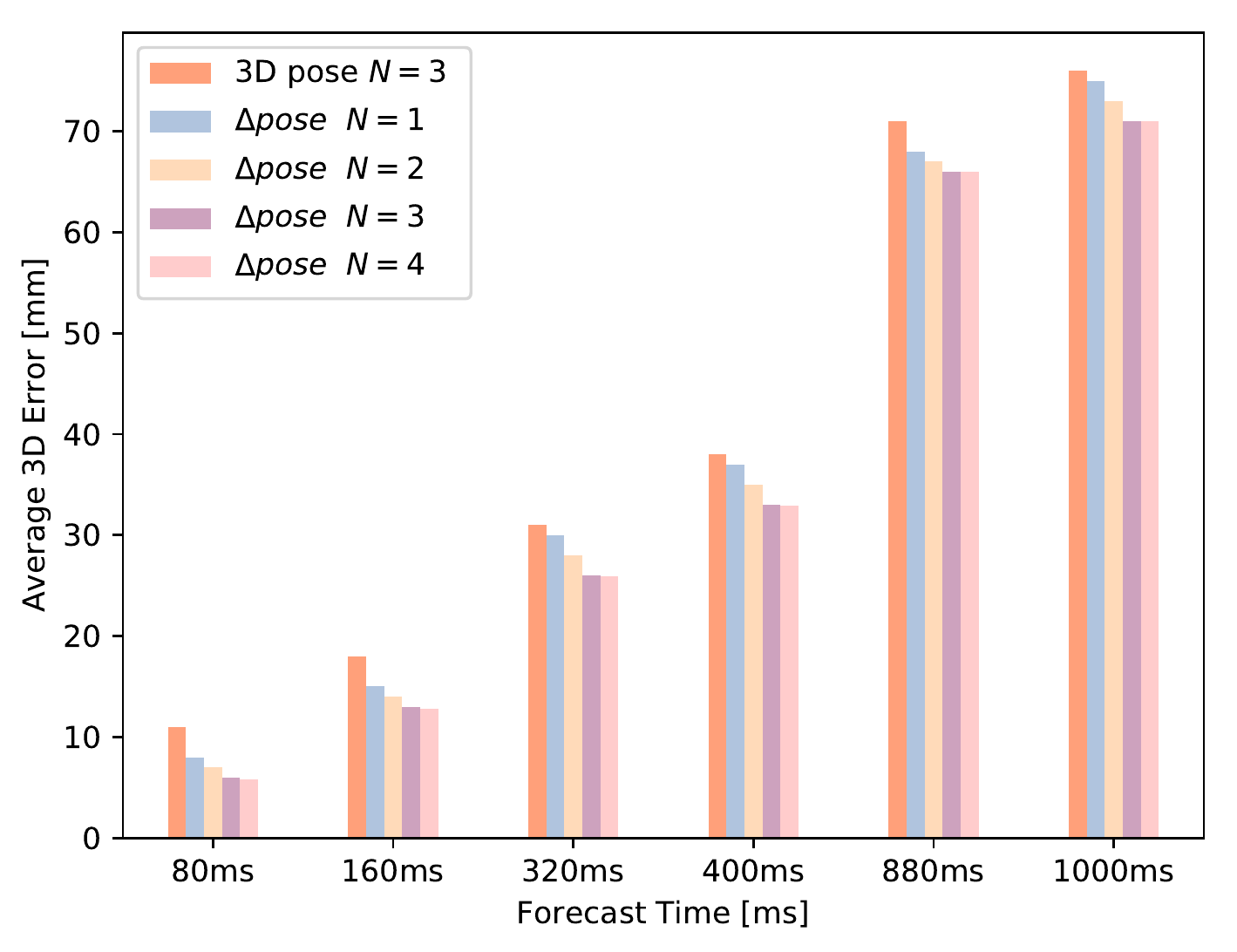}
    \caption{Comparison of average 3D error in $mm$ over all actions of the Human3.6M dataset at different prediction times.}
    \label{fig:ablation}
\end{figure}

\paragraph{Model Architecture.}
We first study the influence of individual components of the proposed method through different ablation studies. Specifically, we report the impact of the number of layers $N$ on the motion prediction. In Fig.~\ref{fig:ablation}, we show the average prediction errors at different forecast times on the Human3.6M dataset. \textit{MotionMixer} yields the best average error with $N = 3$. Stacking more than three layers did not empirically improve the performance. To verify the impact of predicting the pose displacement instead of the 3D body pose, we train our model directly with 3D joints. With a $5mm$ performance gain, our model takes advantage of the pose displacement representation. This is reasonable, since such transformation may help the network focus more on motion patterns rather than the appearance of the body pose, hence, generalizing better to new environments and subjects.

\paragraph{Spatial-Temporal Mixing.} To demonstrate the effect of the \textit{STMixer}, we train the \textit{Spatial-Mix MLP} and the \textit{Temporal-Mix MLP} independently for short-term and long-term prediction and report the results in Tab. \ref{tab:ablation_stmixer}. By removing the temporal or the spatial mixing, the error increases at 1000ms by $4\%$ and $6\%$, respectively. The best results are achieved when simultaneously mixing the body pose in time and space. This is expected since the human spatial-temporal dynamics are interleaved. We also empirically evaluate the effect of the squeeze-and-excitation (SE) blocks, which shed new light on LSTMs hidden-state weighting by giving higher importance to influential time steps. With a 2mm performance gain over all time horizons, the SE-blocks help the network re-calibrate the influence of each pose in the sequence and predict more accurate motion patterns.

\begin{table}[htp!]
\centering
\resizebox{0.48\textwidth}{!}{                                                            
\begin{tabular}{|c|cccccccc|}
& \multicolumn{8}{c|}{Human3.6M}  \\
milliseconds   & 80   & 160  & 320  & 400  & 560 & 720 & 880 & 1000  \\\hline
Spatial-Mix MLP & 12.1 & 17.5 & 32.6 & 37.4 & 51.6 & 61.9 & 70.1 & 77.9  \\
Temporal-Mix MLP & 10.5 & 15.2 & 30.5 & 36.8 & 49.5 & 59.3 & 68.6 & 74.0 \\
STMixer w/o SE-Block  & 8.5 & 14.5 & 29.1 & 35.5 & 48.3 & 58.6 & 67.8 & 73.2  \\
\hline
STMixer  &  \textbf{6.9} & \textbf{12.2} & \textbf{26.9} & \textbf{33.6} & \textbf{46.1} & \textbf{56.5} & \textbf{65.7} & \textbf{71.6}  \\\hline
\end{tabular}
}
\caption{Influence of different parts of the \textit{STMixer} on the performance. ”SE-Block” denotes the squeeze-and-excitation blocks. The best results are shown in bold.}
\label{tab:ablation_stmixer}
\end{table}

\paragraph{Computational Complexity.} We also evaluate the trade-off between the model’s computational cost and performance. The results are shown in Tab.~\ref{tbl:complexity}. We report the number of parameters and an estimate of the floating operations FLOPs to predict 25 frames (1000ms). We compare our model with the current best approaches. In comparison to \cite{mao2021multi}, \textit{MotionMixer} reaches nearly the same performance with only $1.4\%$ of the parameters. We outperform \cite{sofianos2021space} and \cite{dang2021msr} respectively by $4\%$ and $3\%$, while using only $40\%$ and $0.5\%$ of the parameters.  

\begin{table}[h]

	\centering
	\small
	\tabcolsep=1mm
	\resizebox{0.9\linewidth}{!}{
		\begin{tabular}{|c|c|c|c|} 
		Model & Parameters & $\approx$ FLOPs & Average 3D  \\
		\hline
		Motion-Attention \cite{mao2021multi} & 3.4M & - & 110.1 \\
		MSRGCN \cite{dang2021msr} & 6.3M & 192.4M & 114.2 \\
		STSGCN \cite{sofianos2021space} $\dagger$ & 57.5k & 7.1M & 75.6 \\
		\hline
		Ours $N = 1$ & 12.2k & 1.5M & 117.3 \\ 
		Ours $N = 2$ & 18.2k & 1.8M & 115.5 \\
		Ours $N = 3$ & 30.2k & 2.1M & 111.0 \\
		\hline
		Ours  $N = 1$ $\dagger$ & 12.2k & 1.5M & 75.6 \\ 
		Ours $N = 2$ $\dagger$ & 18.2k & 1.8M & 74.8 \\
		Ours  $N = 3$ $\dagger$ & 30.2k & 2.1M & 71.6 \\
		\hline
		\end{tabular}
	}
	\caption{Computational complexity analysis. ($\dagger$) indicates results with the average error over all frames. }
	\label{tbl:complexity}
\end{table}

 \subsection{Limitations}
 
In addition to the qualitative results in Fig.\ref{fig:teaser}, we examine some failure cases of \textit{MotionMixer}. Fig. 5 illustrates an example of the predicted skeletons for the \textit{WalkDog} action. As can be seen, the last three frames do not match the ground-truth poses. This failure is also common for previous methods \cite{mao2021multi,sofianos2021space} since various actions in Human3.6M are performed in different arts in the training dataset. In addition, the human motion is highly uncertain. A sequence of past poses may imply various possible futures. Thus, predicting the inter-joint and inter-frame dependencies in long-term becomes even more complex.

\begin{figure} [htp!]
    \centering
    \vspace{-0.1in}
    \includegraphics[width=0.48\textwidth]{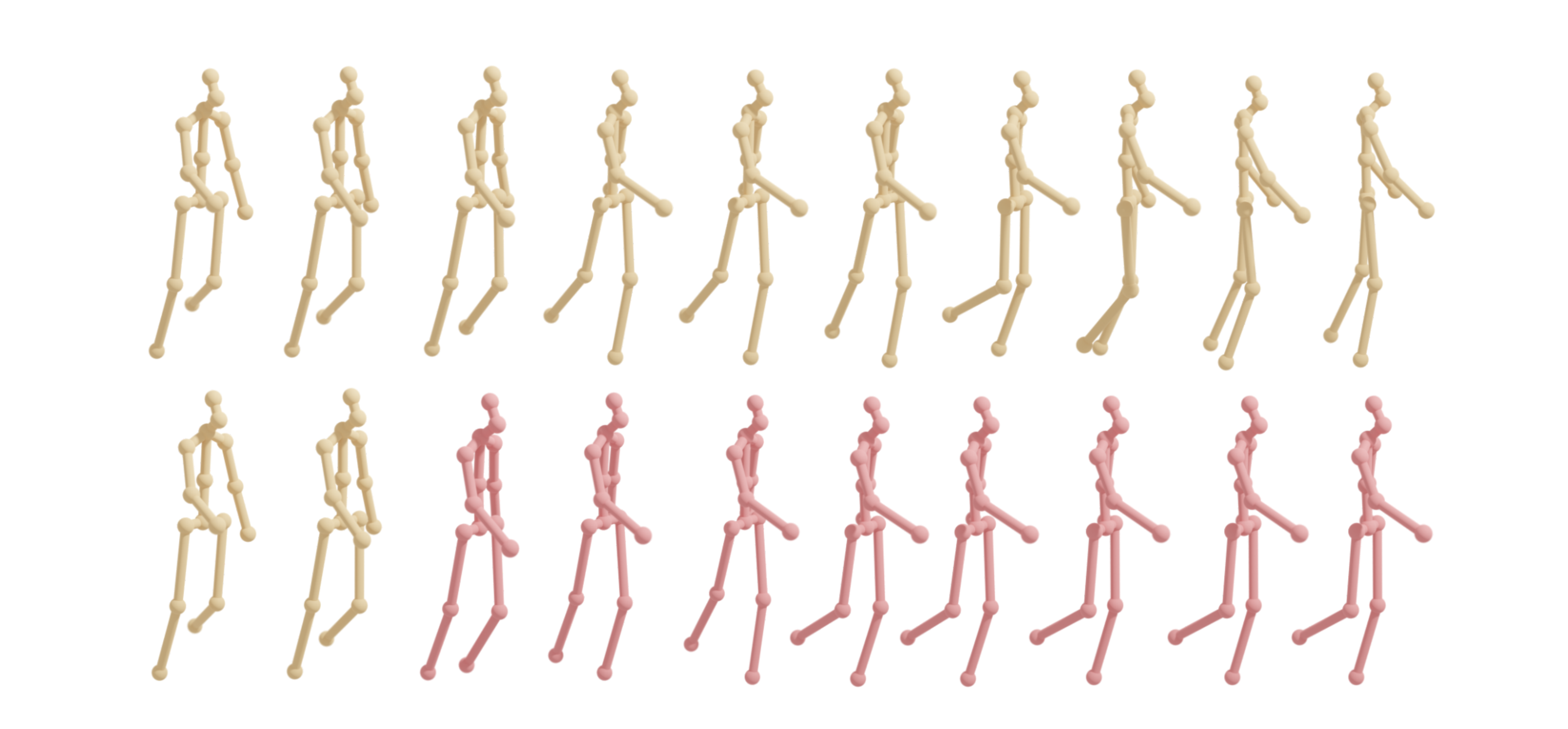}
    \caption{Example of the failure cases of our 3D pose forecasting approach. The first line indicates the ground-truth 3D human motion.  The frames on the left are the observations. The right part, shown in pink is the predicted future motion.}
   \label{fig:failures}
\end{figure}

\section {Conclusion}

In this work, we presented an MLP-based pose forecasting approach that effectively exploits the spatial-temporal dependencies of the 3D human body pose. By learning to mix features across the spatial and temporal domains, our method improved the state-of-the-art for short-term and long-term forecasting on three large-scale benchmark datasets. Enhanced by
squeeze-and-excitation (SE) blocks, which aim to calibrate the influence of each time step in the pose sequence, our model has much less parameters than current best-performing approaches.

\section*{Acknowledgments}
The research leading to these results is funded by the German Federal Ministry for Economic Affairs and Energy within the project “KI Delta Learning" (Förderkennzeichen 19A19013A). The authors would like to thank the consortium for the successful cooperation.
\bibliographystyle{named}
\bibliography{ijcai22}

\begin{thebibliography}{}

\bibitem[\protect\citeauthoryear{Aksan \bgroup \em et al.\egroup
  }{2020}]{aksan2020spatio}
Emre Aksan, Peng Cao, Manuel Kaufmann, and Otmar Hilliges.
\newblock A spatio-temporal transformer for 3d human motion prediction.
\newblock {\em arXiv preprint arXiv:2004.08692}, 2020.

\bibitem[\protect\citeauthoryear{Ba \bgroup \em et al.\egroup
  }{2016}]{ba2016layer}
Jimmy~Lei Ba, Jamie~Ryan Kiros, and Geoffrey~E Hinton.
\newblock Layer normalization.
\newblock {\em arXiv preprint arXiv:1607.06450}, 2016.

\bibitem[\protect\citeauthoryear{Belagiannis \bgroup \em et al.\egroup
  }{2014}]{belagiannis2014holistic}
Vasileios Belagiannis, Christian Amann, Nassir Navab, and Slobodan Ilic.
\newblock Holistic human pose estimation with regression forests.
\newblock In {\em International Conference on Articulated Motion and Deformable
  Objects}, pages 20--30. Springer, 2014.

\bibitem[\protect\citeauthoryear{Bouazizi \bgroup \em et al.\egroup
  }{2021}]{bouazizi2021self}
Arij Bouazizi, Julian Wiederer, Ulrich Kressel, and Vasileios Belagiannis.
\newblock Self-supervised 3d human pose estimation with multiple-view geometry.
\newblock In {\em 2021 16th IEEE International Conference on Automatic Face and
  Gesture Recognition (FG 2021)}, pages 1--8. IEEE, 2021.

\bibitem[\protect\citeauthoryear{Dang \bgroup \em et al.\egroup
  }{2021}]{dang2021msr}
Lingwei Dang, Yongwei Nie, Chengjiang Long, Qing Zhang, and Guiqing Li.
\newblock Msr-gcn: Multi-scale residual graph convolution networks for human
  motion prediction.
\newblock In {\em Proceedings of the IEEE/CVF International Conference on
  Computer Vision}, pages 11467--11476, 2021.

\bibitem[\protect\citeauthoryear{Fragkiadaki \bgroup \em et al.\egroup
  }{2015}]{fragkiadaki2015recurrent}
Katerina Fragkiadaki, Sergey Levine, Panna Felsen, and Jitendra Malik.
\newblock Recurrent network models for human dynamics.
\newblock In {\em Proceedings of the IEEE International Conference on Computer
  Vision}, pages 4346--4354, 2015.

\bibitem[\protect\citeauthoryear{Gui \bgroup \em et al.\egroup
  }{2018}]{gui2018teaching}
Liang-Yan Gui, Kevin Zhang, Yu-Xiong Wang, Xiaodan Liang, Jos{\'e}~MF Moura,
  and Manuela Veloso.
\newblock Teaching robots to predict human motion.
\newblock In {\em 2018 IEEE/RSJ International Conference on Intelligent Robots
  and Systems (IROS)}, pages 562--567. IEEE, 2018.

\bibitem[\protect\citeauthoryear{Hendrycks and
  Gimpel}{2016}]{hendrycks2016gaussian}
Dan Hendrycks and Kevin Gimpel.
\newblock Gaussian error linear units (gelus).
\newblock {\em arXiv preprint arXiv:1606.08415}, 2016.

\bibitem[\protect\citeauthoryear{Hu \bgroup \em et al.\egroup
  }{2018}]{hu2018squeeze}
Jie Hu, Li~Shen, and Gang Sun.
\newblock Squeeze-and-excitation networks.
\newblock In {\em Proceedings of the IEEE conference on computer vision and
  pattern recognition}, pages 7132--7141, 2018.

\bibitem[\protect\citeauthoryear{Ionescu \bgroup \em et al.\egroup
  }{2013}]{ionescu2013human3}
Catalin Ionescu, Dragos Papava, Vlad Olaru, and Cristian Sminchisescu.
\newblock Human3. 6m: Large scale datasets and predictive methods for 3d human
  sensing in natural environments.
\newblock {\em IEEE transactions on pattern analysis and machine intelligence},
  36(7):1325--1339, 2013.

\bibitem[\protect\citeauthoryear{Kingma and Ba}{2014}]{kingma2014adam}
Diederik~P Kingma and Jimmy Ba.
\newblock Adam: A method for stochastic optimization.
\newblock {\em arXiv preprint arXiv:1412.6980}, 2014.

\bibitem[\protect\citeauthoryear{Kuli{\'c} \bgroup \em et al.\egroup
  }{2012}]{kulic2012incremental}
Dana Kuli{\'c}, Christian Ott, Dongheui Lee, Junichi Ishikawa, and Yoshihiko
  Nakamura.
\newblock Incremental learning of full body motion primitives and their
  sequencing through human motion observation.
\newblock {\em The International Journal of Robotics Research}, 31(3):330--345,
  2012.

\bibitem[\protect\citeauthoryear{Li \bgroup \em et al.\egroup
  }{2018}]{li2018convolutional}
Chen Li, Zhen Zhang, Wee~Sun Lee, and Gim~Hee Lee.
\newblock Convolutional sequence to sequence model for human dynamics.
\newblock In {\em Proceedings of the IEEE Conference on Computer Vision and
  Pattern Recognition}, pages 5226--5234, 2018.

\bibitem[\protect\citeauthoryear{Liu \bgroup \em et al.\egroup
  }{2021}]{liu2021motion}
Zhenguang Liu, Pengxiang Su, Shuang Wu, Xuanjing Shen, Haipeng Chen, Yanbin
  Hao, and Meng Wang.
\newblock Motion prediction using trajectory cues.
\newblock In {\em Proceedings of the IEEE/CVF International Conference on
  Computer Vision}, pages 13299--13308, 2021.

\bibitem[\protect\citeauthoryear{Mahmood \bgroup \em et al.\egroup
  }{2019}]{mahmood2019amass}
Naureen Mahmood, Nima Ghorbani, Nikolaus~F Troje, Gerard Pons-Moll, and
  Michael~J Black.
\newblock Amass: Archive of motion capture as surface shapes.
\newblock In {\em Proceedings of the IEEE/CVF International Conference on
  Computer Vision}, pages 5442--5451, 2019.

\bibitem[\protect\citeauthoryear{Mao \bgroup \em et al.\egroup
  }{2019}]{mao2019learning}
Wei Mao, Miaomiao Liu, Mathieu Salzmann, and Hongdong Li.
\newblock Learning trajectory dependencies for human motion prediction.
\newblock In {\em Proceedings of the IEEE/CVF International Conference on
  Computer Vision}, pages 9489--9497, 2019.

\bibitem[\protect\citeauthoryear{Mao \bgroup \em et al.\egroup
  }{2020}]{mao2020history}
Wei Mao, Miaomiao Liu, and Mathieu Salzmann.
\newblock History repeats itself: Human motion prediction via motion attention.
\newblock In {\em European Conference on Computer Vision}, pages 474--489.
  Springer, 2020.

\bibitem[\protect\citeauthoryear{Mao \bgroup \em et al.\egroup
  }{2021}]{mao2021multi}
Wei Mao, Miaomiao Liu, Mathieu Salzmann, and Hongdong Li.
\newblock Multi-level motion attention for human motion prediction.
\newblock {\em International Journal of Computer Vision}, pages 1--23, 2021.

\bibitem[\protect\citeauthoryear{Martinez \bgroup \em et al.\egroup
  }{2017}]{martinez2017human}
Julieta Martinez, Michael~J Black, and Javier Romero.
\newblock On human motion prediction using recurrent neural networks.
\newblock In {\em Proceedings of the IEEE Conference on Computer Vision and
  Pattern Recognition}, pages 2891--2900, 2017.

\bibitem[\protect\citeauthoryear{Sofianos \bgroup \em et al.\egroup
  }{2021}]{sofianos2021space}
Theodoros Sofianos, Alessio Sampieri, Luca Franco, and Fabio Galasso.
\newblock Space-time-separable graph convolutional network for pose
  forecasting.
\newblock In {\em Proceedings of the IEEE/CVF International Conference on
  Computer Vision}, pages 11209--11218, 2021.

\bibitem[\protect\citeauthoryear{Tang \bgroup \em et al.\egroup
  }{2018}]{tang2018long}
Yongyi Tang, Lin Ma, Wei Liu, and Wei-Shi Zheng.
\newblock Long-term human motion prediction by modeling motion context and
  enhancing motion dynamics.
\newblock In {\em Proceedings of the Twenty-Seventh International Joint
  Conference on Artificial Intelligence, {IJCAI-18}}, pages 935--941.
  International Joint Conferences on Artificial Intelligence Organization, 7
  2018.

\bibitem[\protect\citeauthoryear{Tolstikhin \bgroup \em et al.\egroup
  }{2021}]{tolstikhin2021mlp}
Ilya Tolstikhin, Neil Houlsby, Alexander Kolesnikov, Lucas Beyer, Xiaohua Zhai,
  Thomas Unterthiner, Jessica Yung, Andreas~Peter Steiner, Daniel Keysers,
  Jakob Uszkoreit, et~al.
\newblock Mlp-mixer: An all-mlp architecture for vision.
\newblock In {\em Thirty-Fifth Conference on Neural Information Processing
  Systems}, 2021.

\bibitem[\protect\citeauthoryear{Vaswani \bgroup \em et al.\egroup
  }{2017}]{vaswani2017attention}
Ashish Vaswani, Noam Shazeer, Niki Parmar, Jakob Uszkoreit, Llion Jones,
  Aidan~N Gomez, {\L}ukasz Kaiser, and Illia Polosukhin.
\newblock Attention is all you need.
\newblock In {\em Advances in neural information processing systems}, pages
  5998--6008, 2017.

\bibitem[\protect\citeauthoryear{von Marcard \bgroup \em et al.\egroup
  }{2018}]{von2018recovering}
Timo von Marcard, Roberto Henschel, Michael~J Black, Bodo Rosenhahn, and Gerard
  Pons-Moll.
\newblock Recovering accurate 3d human pose in the wild using imus and a moving
  camera.
\newblock In {\em Proceedings of the European Conference on Computer Vision
  (ECCV)}, pages 601--617, 2018.

\bibitem[\protect\citeauthoryear{Wang \bgroup \em et al.\egroup
  }{2007}]{wang2007gaussian}
Jack~M Wang, David~J Fleet, and Aaron Hertzmann.
\newblock Gaussian process dynamical models for human motion.
\newblock {\em IEEE transactions on pattern analysis and machine intelligence},
  30(2):283--298, 2007.

\bibitem[\protect\citeauthoryear{Wiederer \bgroup \em et al.\egroup
  }{2020}]{wiederer2020traffic}
Julian Wiederer, Arij Bouazizi, Ulrich Kressel, and Vasileios Belagiannis.
\newblock Traffic control gesture recognition for autonomous vehicles.
\newblock In {\em 2020 IEEE/RSJ International Conference on Intelligent Robots
  and Systems (IROS)}, pages 10676--10683. IEEE, 2020.

\bibitem[\protect\citeauthoryear{Zhong \bgroup \em et al.\egroup
  }{2022}]{zhong2022spatial}
Chongyang Zhong, Lei Hu, Zihao Zhang, Yongjing Ye, and Shihong Xia.
\newblock Spatial-temporal gating-adjacency gcn for human motion prediction.
\newblock {\em arXiv preprint arXiv:2203.01474}, 2022.

\end{thebibliography}

\end{document}